\documentclass[letterpaper, 10 pt, conference]{ieeeconf}
\IEEEoverridecommandlockouts

\usepackage[utf8]{inputenc} 
\usepackage[T1]{fontenc}    
\usepackage{cite}
\makeatletter
\let\NAT@parse\undefined
\makeatother
\usepackage{hyperref}
\usepackage{url}            
\usepackage{booktabs}       
\usepackage{amsfonts}       
\usepackage{nicefrac}       
\usepackage{microtype}      
\usepackage{xcolor}         
\usepackage{amsmath}
\usepackage{amssymb}
\usepackage{logicpuzzle}
\usepackage{stmaryrd}
\usepackage{algorithmic}
\usepackage{algorithm}
\usepackage{makecell}

\usepackage{amsmath,amsfonts}
\usepackage{algorithmic}
\usepackage{algorithm}
\usepackage{array}
\usepackage[caption=false, font=footnotesize]{subfig}
\usepackage{textcomp}
\usepackage{stfloats}
\usepackage{url}
\usepackage{verbatim}
\usepackage{graphicx}
\usepackage{epstopdf}
\usepackage[utf8]{inputenc}
\usepackage[T1]{fontenc}
\usepackage{booktabs} 
\usepackage{diagbox}  
\usepackage{multirow} 
\usepackage{threeparttable} 
\usepackage{epstopdf} 
\usepackage{colortbl}  
\usepackage{hhline}
\usepackage[inkscapelatex=false]{svg}
\definecolor{codegreen}{rgb}{0,0.5,0}
\definecolor{codeblue}{rgb}{0,0,0.9}
\definecolor{codeblues}{rgb}{0,0,0.4}
\definecolor{codegray2}{rgb}{0.4,0.4,0.4}
\definecolor{codegray}{rgb}{0.9,0.9,0.9}

\usepackage{pifont}

\usepackage{color, xcolor}

\newcounter{ALC@tempcntr}

\usepackage{fontawesome}
\definecolor{Dred}{HTML}{C81D31}
\definecolor{Dgreen}{HTML}{588E31}
\definecolor{codegray}{rgb}{0.9,0.9,0.9}

\begin{document}
\title{SCOPE: Scene-graph Construction Optimized by Plan Error-correction for Robot Long-term Planning Datasets}

\author{Chaoyue Li$^{1,2,*}$, Zeyu Gao$^{1,2,*}$, Yanfeng Lu$^{1,2,\dagger}$, Fan Luo$^{1,2}$%
\thanks{$^{1}$State Key Laboratory of Multimodal Artificial Intelligence Systems,
Institute of Automation, Chinese Academy of Sciences (CASIA), Beijing 100190, China.}%
\thanks{$^{2}$School of Artificial Intelligence, University of Chinese Academy of Sciences (UCAS), Beijing 100049, China.}%
\thanks{$^{*}$These authors contributed equally.}%
\thanks{$^{\dagger}$Correspondence to: Yanfeng Lu \texttt{\textless{}yanfeng.lv@ia.ac.cn\textgreater{}}.}%
}

\maketitle
\pagestyle{empty}
\thispagestyle{empty}

\begin{abstract}
Task and Motion Planning (TAMP) stands as a fundamental challenge in robotics, essential for enabling autonomous agents to execute complex, long-horizon tasks in diverse environments. While Large Language Models (LLMs) introduce a new paradigm for TAMP by leveraging extensive common sense, their inherent reasoning limitations hinder direct application, and fine-tuning strategies to enhance these capabilities are currently bottlenecked by the scarcity of high-quality datasets. Existing automated generation methods often suffer from homogeneity, limiting model generalization, and unsolvability, where hallucinated scene graphs create physical mismatches with task goals. To address these issues, we propose \textbf{S}cene-graph \textbf{C}onstruction \textbf{O}ptimized by \textbf{P}lan \textbf{E}rror-correction (\textbf{SCOPE}), an automated framework designed to generate diverse and executable TAMP datasets. SCOPE incorporates multi-level diversity metrics to maximize the information entropy of scenarios. Crucially, it introduces a Plan Feedback Error-correction pipeline that utilizes planner execution failures as feedback signals to retroactively rectify logical inconsistencies in scene graphs, ensuring physical feasibility. Experimental results demonstrate that SCOPE exhibits greater robustness to erroneous scene graphs and generates higher-quality data. When utilized for LLM fine-tuning, SCOPE-generated data yield a downstream TAMP model that achieves an absolute 12.33\% gain in success rate over its counterpart fine-tuned with data generated by DELTA, a state-of-the-art PDDL-based planner.
More demos and information are available on \href{https://scope-page.github.io/}{https://scope-page.github.io/}.
\end{abstract}


\section{Introduction}

Task and Motion Planning (TAMP) stands as a fundamental challenge in the field of robotics. Traditional approaches~\cite{chitnis2016guided, jiang2019task,akbari2019combined} rely primarily on rule-based designs and heuristic searches. However, these methods often lack sufficient generalization capabilities. The emergence of Large Language Models (LLMs)~\cite{achiam2023gpt,bai2023qwen} has introduced a new paradigm~\cite{song2023llm, liu2023llm} for addressing TAMP challenges. By leveraging their extensive common sense knowledge, LLMs demonstrate significant potential in handling complex TAMP tasks in different scenarios, enhancing the universality of planning systems.

Current methods utilizing LLMs for TAMP can be broadly categorized into Symbolic Planning and Direct Planning. Symbolic Planning~\cite{rana2023sayplan,mandi2024roco,zhang2025lamma} employs the LLM as a translator that decomposes tasks based on hand-crafted rules to generate problem definitions for traditional planners. While this approach achieves a relatively high success rate, it relies heavily on manually defined domains and rules. Consequently, this paradigm exhibits limited error tolerance, making full automation difficult. Conversely, the Direct Planning methods~\cite{song2023llm,joublin2024copal,pchelintsev2025lera} utilize the LLM directly as a planner. These methods exploit the common sense reasoning capabilities of the model and require minimal manual prior constraints, allowing users to accomplish a wide range of tasks through natural language. Nevertheless, the success rate of this approach is often limited by the inherent shortcomings of LLMs in temporal reasoning and their tendency towards hallucination. Ultimately, the performance of both paradigms is intrinsically constrained by the underlying planning capabilities of the LLMs.
\begin{figure}[t]
    \centering
    \includegraphics[width=\columnwidth]{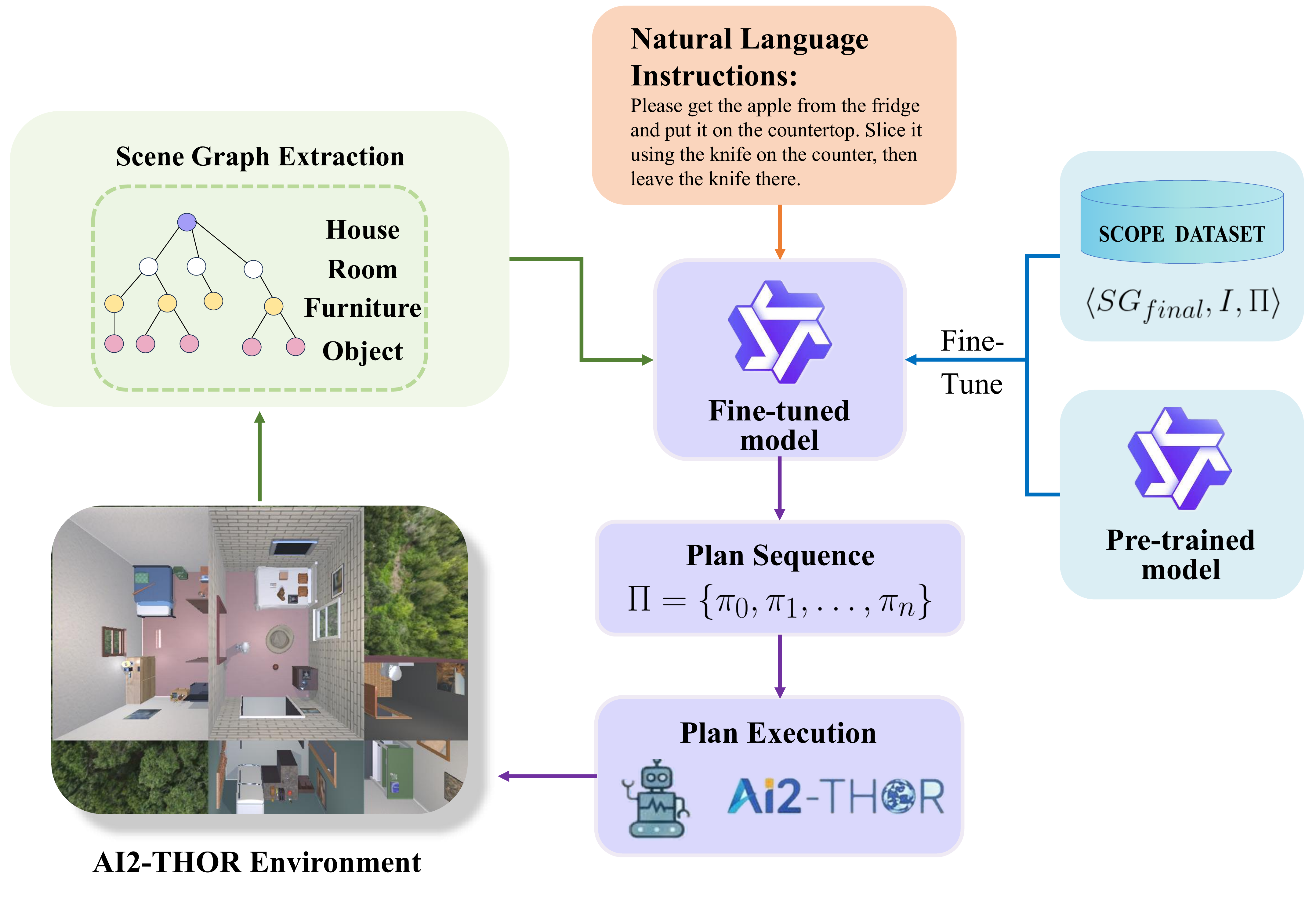}
    \vspace{-15pt}
    \caption{An end-to-end task planning framework demonstrates the complete pipeline from environmental perception to plan generation. First, hierarchical scene graphs are extracted from the AI2-THOR environment. Subsequently, these graphs and natural language instructions are fed into the Qwen2.5-7B-Instruct model. This model is fine-tuned on the SCOPE dataset via LoRA for efficient task planning. Finally, the model outputs an ordered plan sequence $\Pi = \{\pi_0, \pi_1, \dots, \pi_n\}$, which is executed by the AI2-THOR simulator.}
    \vspace{-12pt}
    \label{fig:finetune}
\end{figure}
Therefore, fine-tuning LLMs with extensive data serves as an effective strategy~\cite{mu2024robocodex, mu2023embodiedgpt} to internalize planning logic and enhance temporal reasoning capabilities. A standard TAMP fine-tuning sample consists of three core elements: the scene graph describing the environmental state, the instruction defining the task goal, and the plan serving as the solution path. However, constructing TAMP datasets currently requires balancing data quality and annotation cost. Manually annotated datasets~\cite{yuan2024tasklama} offer high quality yet incur prohibitive costs, making them difficult to scale for fine-tuning needs. While data constructed using LLMs~\cite{valmeekam2023planbench} offers a low-cost automated alternative, it often suffers from low quality. 

This deficiency primarily manifests in two aspects. First, the generated content suffers from a lack of diversity. Due to the preference of LLMs, the scene graphs often exhibit high repetitiveness, resulting in insufficient information entropy within the dataset and limiting the generalization ability of the model. Second, the generated tasks frequently suffer from unsolvability because of the wrong scene graphs. Influenced by hallucinations, automatically generated scene graphs often contain logical flaws, such as missing key interactive objects or incorrect attributes. This mismatch between the environment and the task renders the generated problems physically unsolvable, thereby introducing a significant amount of noise into the training data.

To address these challenges, we propose \textbf{S}cene-graph \textbf{C}onstruction \textbf{O}ptimized by \textbf{P}lan \textbf{E}rror-correction (\textbf{SCOPE}), an automated generation framework designed specifically for constructing high-quality TAMP datasets. To mitigate the homogeneity inherent in LLM-generated content, SCOPE incorporates multi-level Diversity Metrics during the Scene Graph Construction phases. Specifically, we monitor Global Diversity to ensure a broad coverage of task types while leveraging Local Diversity to enrich fine-grained variations in object attributes and spatial relationships, thereby maximizing the information entropy of the dataset. Second, to tackle the prevalent issue of unsolvability in automatically generated data, SCOPE introduces a Plan Feedback Error-correction pipeline. This pipeline establishes a closed-loop system for generation and verification, where execution results from the planner serve as feedback signals. This process allows the system to retroactively identify and correct logical inconsistencies within the scene graph while iteratively generating feasible planning sequences, effectively solving mismatches between the environment and the task.

Our main contributions are summarized as follows:
\begin{itemize}
\item We propose SCOPE, an automated data generation framework that integrates diversity-aware generation and Plan Feedback Error-correction. By maximizing data diversity and iteratively rectifying logical inconsistencies through execution feedback, SCOPE effectively constructs highly diverse and physically executable TAMP datasets with high quality.
\item We investigate the impact of data quality on the end-to-end planning performance of fine-tuned LLMs through extensive experiments, as illustrated in Fig.~\ref{fig:finetune}. Validated in the AI2-THOR environment, our results demonstrate that the high-quality data generated by SCOPE significantly improves the planning success rate, achieving an absolute 12.33\% gain over data generated by DELTA.
\end{itemize}

\section{Related Work}
\subsection{Task and Motion Planning with LLMs}

With their remarkable capabilities in common sense reasoning and semantic understanding, LLMs have introduced a paradigm shift in solving TAMP problems. Existing approaches utilizing LLMs can be broadly categorized into two distinct streams: Symbolic Planning and Direct Planning.

In the Symbolic Planning paradigm, the LLM functions primarily as a semantic translator to bridge natural language and classical solvers. LLM+P~\cite{liu2023llm} pioneered this approach by converting natural language instructions into PDDL problem definitions to leverage the optimality of external planners. To scale this approach to large environments, SayPlan~\cite{rana2023sayplan} grounds LLMs into 3D scene graphs, utilizing semantic search to extract relevant sub-graphs and reduce the problem scope. To address long-horizon tasks, LLMGenPlan~\cite{silver2024generalized} investigates generalized planning, in which the LLM synthesizes generic Python policy programs for an entire class of problems within a domain and refines them through an automated debugging loop. DELTA~\cite{liu2025DELTA} integrates LLM-based scene graph pruning and goal decomposition, breaking down complex objectives into sequences of solvable sub-problems. 

Conversely, the Direct Planning paradigm utilizes the LLM directly as a planner. LLM-As-Planner~\cite{huang2022language} demonstrates the zero-shot capability of models to generate high-level action sequences based on common sense knowledge. RoCo~\cite{mandi2024roco} facilitates collaborative planning through dialog. CoPal~\cite{joublin2024copal} incorporates iterative feedback mechanisms to correct execution failures. DAG-Plan~\cite{gao2024dag} proposes generating Directed Acyclic Graphs to explicitly model temporal dependencies.

Ultimately, the performance of both paradigms is intrinsically constrained by the underlying planning capabilities of the base models.

\subsection{Instruction Tuning and Automated Data Generation}

Instruction tuning has emerged as a pivotal paradigm for unlocking the complex reasoning and planning capabilities of LLMs~\cite{mu2024robocodex, yuan2024tasklama}. However, the efficacy of this approach is intrinsically tied to the quality and diversity of the training corpora~\cite{longpre2023flan}. While foundational datasets such as TaskLama~\cite{yuan2024tasklama} and FLAN~\cite{longpre2023flan} offer high-quality examples, they rely on expensive human annotation and are inherently limited in scale. To address data scarcity, previous works have attempted to synthesize training data using LLM~\cite{wang2022super}. Although scalable, these methods often introduce noise due to LLM hallucinations, leading to mismatches between instructions and responses. To mitigate this, recent approaches leverage automated PDDL planners~\cite{valmeekam2023planbench,verma2025teaching} to construct instruction tuning datasets, yielding higher logical consistency than purely LLM-generated data. However, existing PDDL-based pipelines typically produce scene graphs that are structurally simple and lack diversity, resulting in homogeneous environment configurations. Furthermore, due to persistent hallucinations in the task formulation process, these methods often fail to guarantee solvability, frequently generating invalid samples where task goals are logically incompatible with the initial environmental states.

\section{Methods}
\begin{figure*}[htpb]
    \vspace{3pt}
    \centering
    \includegraphics[width=\textwidth]{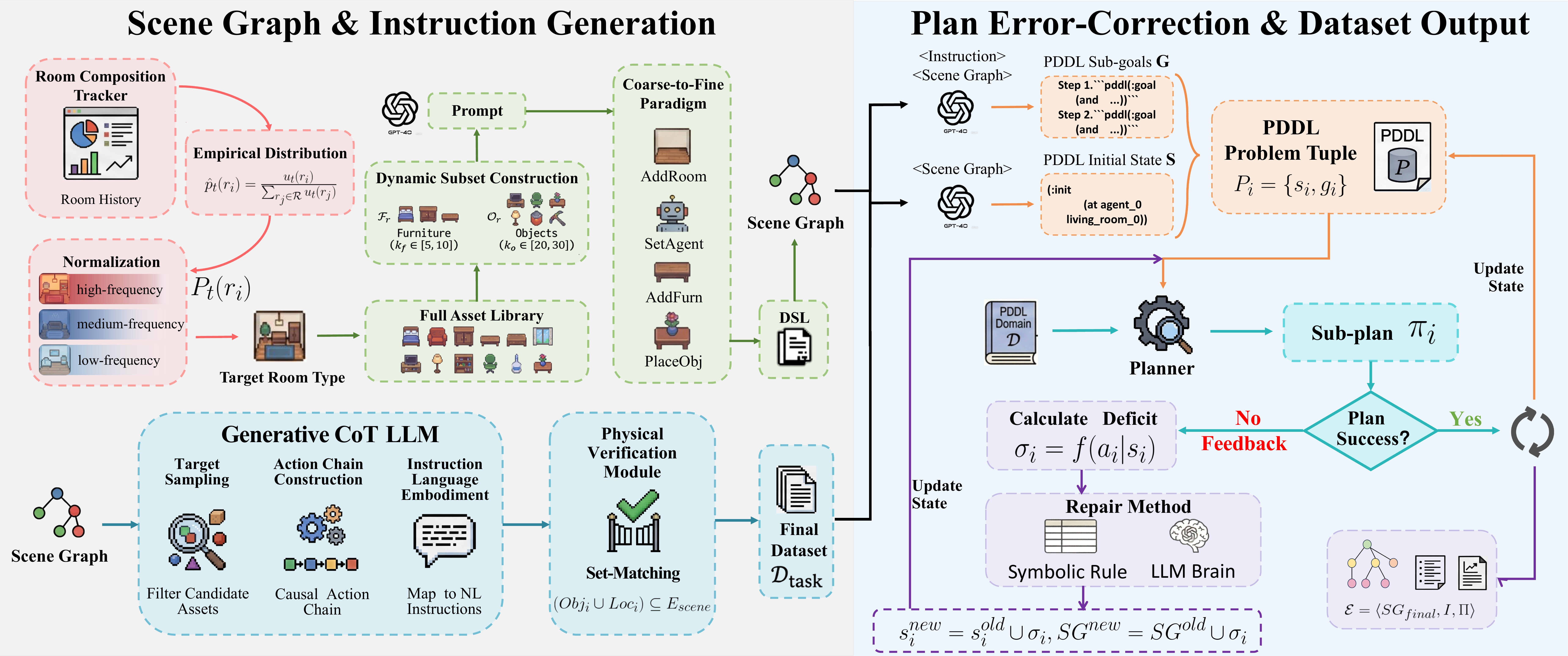}
    \vspace{-10pt}
    \caption{Overview of the SCOPE framework. The pipeline consists of three main stages: (1) diverse scene generation based on statistical feedback and dynamic sampling; (2) Chain-of-Thought driven task instruction generation; and (3) scene graph construction optimized by planning feedback and autoregressive error-correction.}
    \label{fig:framework}
    \vspace{-5pt}
\end{figure*}

To address the prevalent challenges of homogeneity and physical unsolvability inherent in existing automated data generation, we propose SCOPE which operates as a closed-loop generation framework that iteratively refines environment configurations to align with task goals as shown in Fig. \ref{fig:framework}. In this section, we introduce SCOPE through three subsections corresponding to scene graph generation, instruction generation, and plan generation, followed by a comprehensive description of the overall SCOPE framework.
\subsection{Diverse Scene Graph Generation via Statistical Feedback and Dynamic Sampling}
\label{subsec:diversity_generation}

To overcome the LLM's biased preference for high-frequency, conventional room types, we introduce a room composition tracker that continuously monitors the statistical distribution of historically generated scenes. We define the set of room types as $\mathcal{R}$. At generation step $t$, let the cumulative usage frequency of room type $r_i$ be $u_t(r_i)$. The empirical distribution of room types in the generation history can then be expressed as $\hat{p}_t(r_i) = u_t(r_i) / \sum_{r_j \in \mathcal{R}} u_t(r_j)$. The entropy of this distribution, $H_t = -\sum_{r_i \in \mathcal{R}} \hat{p}_t(r_i) \log \hat{p}_t(r_i)$, reflects the diversity of room types in the generated scenes up to step $t$: lower entropy indicates a more concentrated distribution and a tendency toward repetitive generation. Based on this information-theoretic insight, we design an entropy-aware inverse-frequency weighting mechanism that dynamically adjusts the selection probability to maximize future entropy:
\begin{equation}
\label{eq:entropy_sampling}
\begin{aligned}
\hat{p}_t(r_i) &= \frac{u_t(r_i)}{\sum_{r_j \in \mathcal{R}} u_t(r_j)} \\
P_t(r_i) &= \frac{\exp(-\alpha \cdot \hat{p}_t(r_i))}{\sum_{r_j \in \mathcal{R}} \exp(-\alpha \cdot \hat{p}_t(r_j))}
\end{aligned}
\end{equation}
where $\alpha$ is an entropy regularization coefficient (empirically set to $3.0$) that controls the penalty strength applied to frequently selected room types. This formulation establishes a global-level feedback loop grounded in entropy maximization: as a room type $r_i$ becomes overrepresented in the generation history, its empirical probability $\hat{p}_t(r_i)$ increases, which in turn reduces its future selection probability $P_t(r_i)$ through the exponential weighting scheme, thereby compelling the model to explore under-utilized room configurations from the long-tail distribution.

Beyond room-type diversity, we further address the repetitiveness at the asset level. Even when the same room type is selected across multiple generations, LLMs tend to produce similar sets of furniture and objects. To mitigate this, for each target room type $r \in \mathcal{R}$, we randomly sample a restricted subset $\mathcal{M}_{sub}$ from the original asset library $\mathcal{M}$:

\begin{equation}
\begin{aligned}
\mathcal{M}_{sub} = &\text{Sample}(\mathcal{F}_r, k_f) \cup \text{Sample}(\mathcal{O}_r, k_o), \\
&\text{where}\quad \mathcal{F}_r, \mathcal{O}_r \subseteq  \mathcal{M}.
\end{aligned}
\end{equation}
The $\mathcal{F}_r$ and $\mathcal{O}_r$ denote the sets of available furniture and objects for room $r$, respectively, with sampling sizes $k_f \in [5, 10]$ and $k_o \in [20, 30]$. This randomized prompt injection ensures significant differentiation in asset selection across various generation instances, preventing the model from defaulting to the same high-frequency items and encouraging exploration of a wider variety of asset combinations.

In addition to diversity, ensuring the logical correctness of generated scene graphs is paramount. We observe that requiring LLMs to directly output complete scene graphs in a single step frequently leads to structural errors, such as objects placed without supporting furniture or inconsistent containment relationships. To address this, we design a lightweight Domain-Specific Language (DSL) that transforms scene generation into a programmatic instruction stream. The construction process follows a coarse-to-fine paradigm: initially, \texttt{AddRoom} and \texttt{SetAgent} establish the room type and the agent's initial state; subsequently, \texttt{AddFurn} instantiates large furniture structures to define the primary layout of the room; finally, \texttt{PlaceObj} associates fine-grained objects with the interior or exterior of the furniture, completing the macro-to-micro population.

\subsection{Chain-of-Thought Driven Task Instruction Generation}
\label{subsec:instruction_generation}

To construct semantically rich and difficulty-graded instructions for the TAMP dataset $\mathcal{D}_{task}$, we automatically generate them using Chain-of-Thought (CoT) reasoning~\cite{wei2022chain} via LLMs. The pipeline adopts a scene-entity-oriented generation strategy. Leveraging LLM logical reasoning, it directly utilizes the furniture and object nodes $\mathcal{V}_{obj}$ within the scene graph $SG$ to generate physically plausible task instructions $I$.

To guarantee task physical feasibility, we designed a strict CoT logical sequence: Target Sampling $\rightarrow$ Action Chain Construction $\rightarrow$ Instruction Language Embodiment. LLMs first extract object attributes and spatial relations from the linearized scene graph to filter candidate assets with interaction potential. Constrained by embodied configurations and sampled assets, LLMs then construct a complete causal action chain locating the final target state, ensuring logical coherence and physical executability. Finally, the LLM maps this chain into semantically rich natural language task instructions. By varying how explicitly the user intent is expressed, we synthesize descriptions across three difficulty levels: simple, complex, and vague.

To ensure entity consistency in generated instructions, we introduce a set-matching-based verification module during post-processing. This module verifies whether target objects and locations parsed from the instruction are contained in the scene entities (i.e., $(Obj_i \cup Loc_i) \subseteq E_{scene}$). Only tasks whose interactive elements objectively exist in the environment are retained.
\subsection{Plan Generation and Error-Correction}
\label{subsec:planning_feedback_repair}
After obtaining the scene graph and instruction, SCOPE generates an executable plan to complete the dataset. Concretely, SCOPE decomposes the task and plans with a PDDL solver, leveraging the planning process as a diagnostic signal to close the loop between plan generation and scene graph correction: solver failures or inconsistent plans indicate latent logical errors in the scene graph (e.g., missing support/containment relations, invalid initial states, or unreachable target entities).

\subsubsection{Task Decomposition and Formalization}
SCOPE encapsulates low-level robot manipulation capabilities into 9 high-level atomic actions: \texttt{place\_on}, \texttt{place\_in}, \texttt{open}, \texttt{close}, \texttt{power\_on}, \texttt{power\_off}, \texttt{fill}, \texttt{pour}, \texttt{slice}. Given instruction $I$ and initial scene graph $SG_{init}$, we linearize and parse $SG_{init}$ into an initial PDDL state set $s_0$. Simultaneously, leveraging LLM semantic understanding, we progressively decompose the ambiguous natural language instruction $I$ into an action sequence $\mathcal{A} = \{a_0, a_1, \dots, a_n\}$, which LLMs transform into sub-goals $G = \{g_0, g_1, \dots, g_n\}$ based on PDDL domain $\mathcal{D}$.

Through this process, the system constructs a standard PDDL planning problem tuple $P_{i} = \{ s_{i}, g_{i} \}$. The PDDL planner aims to resolve sub-goal $g_i$ given current state $s_i$, returning the subsequent state $s_{i+1}$ after executing plan $\pi_i$.

This procedure rigorously transforms natural language instructions into a machine-verifiable symbolic logical representation.

\subsubsection{Autoregressive Error-Correction and Optimization of Scene Graphs via Planning Feedback}

After task decomposition, complex instructions become serialized sub-task goals. To ensure sub-goal physical executability, our framework introduces a classical heuristic search planner as the core verification and execution engine. Similar to DELTA autoregressive task planning, the planner searches for a viable path from the current state set $s_{i}$ to the target state constrained by the PDDL domain $\mathcal{D}$:

\begin{equation}
    \pi_{i} = f_{plan}(\mathcal{D}, P_{i}).
\end{equation}

If the planner successfully solves a valid sub-plan path $\pi_{i}$, we easily obtain a viable plan from $s_0$ to $s_n$.

However, because LLM-generated scene graphs frequently miss entity attributes due to hallucination (e.g., lacking an openable state, preventing the robot from reaching the target), planning processes easily fall into unsolvable deadlocks. Instead of a passive discard-on-error strategy, SCOPE transforms the unsolvable planner state into a trigger signal for reverse repair. Intercepting this anomaly signal, the system retroactively traces back to the core target action $a_{i}$ of the failed sub-task and analyzes the structural state deficiency $\sigma_{i}$ causing the planning failure:

\begin{equation}
    \sigma_{i} = f(a_{i}|s_{i}),
\label{sigma}
\end{equation}

where $f(a_{i}|s_{i})$ denotes the necessary preconditions for validly executing action $a_{i}$ in state $s_{i}$. Solving for $\sigma_{i}$ concretizes the ambiguous black-box error reported by the planner into precise static physical attributes or initial dynamic states missing from the scene graph. Subsequently, injecting these missing elements into corresponding entity nodes yields the optimized, complete new state and scene graph:

\begin{equation}
\begin{aligned}
    s^{new}_{i} = s^{old}_{i} \cup  \sigma_{i},\\
    SG^{new} = SG^{old} \cup  \sigma_{i}.
\end{aligned}
\end{equation}

This satisfies the prerequisite logical constraints for action execution, allowing the failed planning process to resume.
To determine the state deficiency $\sigma_i$, we construct a Symbolic Rule-LLM dual-layer module as $f$ in Eq.~(4) for scene graphs based on planning feedback. As the foundational refinement layer, the system extracts anomalous objects and predicates, utilizing a predefined action precondition mapping table for precise matching (e.g., adding a ``powerable'' attribute when \texttt{power\_on} planning fails). For long-tail actions missing the rule table, the system jointly feeds restricted object states and the failed action into the LLM. Leveraging LLM robust common-sense reasoning as an advanced semantic supplement, the system reverse-engineers and merges static attributes and initial states required to legitimize the action.

This closed-loop correction process returns the repaired scene state to the planner for verification until a valid sub-plan $\pi_i$ is obtained. The system autoregressively iterates through all sub-tasks, concatenating all valid sub-plans $\pi_{i}$ into a complete task sequence $\Pi=\{\pi_0,\pi_1,\dots,\pi_n\}$. This endows data generation with self-correcting capabilities, ensuring strict alignment between the scene graph and task logic.

This iterative Decomposition-Planning-Repair closed-loop mechanism effectively overcomes model-generated logical hallucinations. Ultimately, it yields a physically executable, high-quality data tuple $\mathcal{E} = \langle SG_{final}, I, \Pi \rangle$ (the corrected scene graph, task instruction, and valid planning path), providing robust training data for fine-tuning LLM end-to-end common-sense planning capabilities.

\subsection{Overview of the SCOPE Framework and Pipeline Design}
\label{subsec:pipeline_overview}

As illustrated in Fig.~\ref{fig:framework}, SCOPE constructs scene-instruction pairs and turns them into executable planning data. During scene generation, statistical feedback and dynamic asset sampling encourage long-tail configurations, while a lightweight Domain-Specific Language (DSL) maps programmatic instructions into the scene graph $SG$. After linearizing $SG$, SCOPE leverages LLM generative CoT reasoning to synthesize task instructions through Target Sampling $\rightarrow$ Action Chain Construction $\rightarrow$ Instruction Language Embodiment, followed by set-matching verification for valid task instructions $I$.

To ensure physical executability, the tuple $\langle SG,I\rangle$ is transformed into serialized PDDL sub-problems for sequential heuristic planning. Upon planning failure, SCOPE estimates the structural state deficiency $\sigma_i$, repairs the scene graph through the Symbolic Rule-LLM module, and replans until valid sub-plans $\pi_i$ are obtained. These sub-plans are concatenated into $\Pi$, yielding the final tuple $\mathcal{E}=\langle SG_{final},I,\Pi\rangle$ and providing robust data for fine-tuning embodied models.

\begin{algorithm}[htbp]
\caption{SCOPE}
\label{alg:scope_pipeline}
\begin{algorithmic}[1]
\REQUIRE Asset Database $\mathcal{M}$, Room Types $\mathcal{R}$, Dataset Size $N$, Domain $\mathcal{D}$, Action Sequence $\mathcal{A}$
\ENSURE High-quality  Dataset $\Omega = \{ \langle SG, I, \Pi \rangle_i \}_{i=1}^N$

\STATE \textbf{Initialize:} $\Omega \leftarrow \emptyset, \; t \leftarrow 0$
\WHILE{$|\Omega| < N$}
    \STATE $t \leftarrow t + 1$
    
    \STATE \textit{// Phase 1: Diverse Scene Generation}
    \STATE Update $P_t(r)$ via entropy Eq.(1)
    \STATE $r \sim P_t(r),\mathcal{M}_{sub} \sim \text{Sample}(\mathcal{M}, r)$
    \STATE $SG \leftarrow \text{Compiler}(\text{LLM}_{scene}(\mathcal{M}_{sub}, r))$
    
    \STATE \textit{// Phase 2: CoT-Driven Instruction Generation}
    \STATE $I \leftarrow \text{LLM}_{cot-task}(SG)$
    
    \IF{$\text{Entities}(I) \not\subseteq \text{Entities}(SG)$}
        \STATE \textbf{continue}
    \ENDIF
   \STATE \textit{// Phase 3: Plan Generation }
    \STATE $\mathcal{A} \leftarrow \text{Decompose}(I, SG_{init})$
    \STATE $G \leftarrow \text{MapToGoals}(\mathcal{A})$
    \STATE $s_{0} \leftarrow \text{ToPDDL}(G_{init}),\Pi \leftarrow \emptyset,i \leftarrow 0$ 
    \FOR{\textbf{each} sub-goal $g_{i} \in \ G$}
        \STATE $\pi_{i} \leftarrow f_{plan}(\mathcal{D}, P_{i})$
        \WHILE{$\pi_{i} = \text{Failure}$}
            \STATE $\sigma_{i}  \leftarrow f(a_{i}|s_{i})$
            \STATE $s^{new}_{i} \leftarrow s^{old}_{i} \cup  \sigma_{i},SG^{new} \leftarrow SG^{old} \cup  \sigma_{i}$
            \STATE $P^{new}_{i}\leftarrow(s^{new}_{i},g_{i}),\pi_{i} \leftarrow f_{plan}(\mathcal{D}, P^{new}_{i})$ 
        \ENDWHILE
        \STATE $\Pi \leftarrow \Pi \cup \pi_{i}$
        \STATE $s_{i+1} \leftarrow \text{Apply}(\pi_{i}, s_{i})$
    \ENDFOR
    \STATE $\ SG_{final}  \leftarrow SG^{new}$
    \STATE $\Omega \leftarrow \Omega \cup \{ \langle SG_{final}, I, \Pi \rangle \}$
\ENDWHILE

\STATE \textbf{return} $\Omega$
\end{algorithmic}
\end{algorithm}

\section{Experiments}
In the experiments, we are interested in answering the following questions about SCOPE:
\begin{itemize}
\item[$\bullet$] How does the SCOPE framework perform in terms of planning robustness, compared to traditional planning algorithms, when confronted with input scene graphs containing logical errors?
\item[$\bullet$] What quality improvements does the data generated by SCOPE offer relative to existing data generation methods? Furthermore, what specific impact do datasets of varying quality have on the downstream fine-tuning efficacy of LLMs for TAMP planning tasks?
\end{itemize}

\subsection{Experimental Setup}
\label{subsec:experimental_setup}

\subsubsection{Dataset Construction}
To address the two core questions, we construct two independent datasets, each containing 100 scene graphs and 300 instructions categorized into three difficulty levels:
\begin{itemize}
    \item \textbf{Simple Tasks}: Feature explicit instructions and short execution paths to evaluate basic action mapping and commonsense object interaction capabilities.
    \item \textbf{Complex Tasks}: Involve long-horizon logic and multi-subtask dependencies, requiring models to disentangle causal relationships and satisfy physical preconditions.
    \item \textbf{Vague Tasks}: Contain semantic ambiguity or abstract intents to verify intent comprehension and reasoning capabilities under uncertain states.
\end{itemize}
Based on this taxonomy, we devise two datasets from distinct sources. The \textbf{LLM-Synthetic Noisy Dataset} comprises LLM-generated scene graphs and instructions whose scene graphs exhibit logical hallucinations and attribute omissions, and is used to evaluate the robustness of planning methods to noisy scene-graph states, as well as the effectiveness of SCOPE's feedback-driven repair mechanism. In contrast, the \textbf{AI2-THOR Grounded Dataset} extracts scene graphs directly from AI2-THOR simulator states and is used to evaluate the fine-tuned embodied model's final planning generalization performance in realistic, logic-flaw-free environments.
\begin{table*}[t]
\vspace{3pt}
\centering
\caption{Planning Robustness Evaluation on LLM-Synthetic Noisy Dataset. Comparison of baseline models and SCOPE under erroneous scene graphs generated by LLMs.}
\vspace{-5pt}
\label{tab:robustness_dataset_a}
\resizebox{\textwidth}{!}{%
\begin{tabular}{lcccccccccccc}
\toprule
\multirow{2}{*}{\textbf{Methods}} & \multicolumn{3}{c}{\textbf{Overall}} & \multicolumn{3}{c}{\textbf{Simple Tasks}} & \multicolumn{3}{c}{\textbf{Complex Tasks}} & \multicolumn{3}{c}{\textbf{Vague Tasks}} \\
\cmidrule(lr){2-4} \cmidrule(lr){5-7} \cmidrule(lr){8-10} \cmidrule(lr){11-13}
 & \textbf{AQS~$\uparrow$} & \textbf{SR~$\uparrow$} & \textbf{EX~$\uparrow$} & \textbf{AQS~$\uparrow$} & \textbf{SR~$\uparrow$} & \textbf{EX~$\uparrow$} & \textbf{AQS~$\uparrow$} & \textbf{SR~$\uparrow$} & \textbf{EX~$\uparrow$} & \textbf{AQS~$\uparrow$} & \textbf{SR~$\uparrow$} & \textbf{EX~$\uparrow$} \\
\midrule
LLM-As-Planner~\cite{song2023llm} & 1.84 & 27.00\% & 29.67\% & 2.27 & 46.00\% & 46.00\% & 1.78 & 18.00\% & 21.00\% & 1.47 & 17.00\% & 22.00\% \\
LLM+P~\cite{liu2023llm} & 1.76 & 41.67\% & 48.00\% & 2.68 & 67.00\% & 67.00\% & 1.41 & 30.00\% & 39.00\% & 1.18 & 28.00\% & 38.00\% \\
LLMGenPlan (w/o rp.) & 0.40 & 6.33\% & 36.67\% & 0.61 & 12.00\% & 45.00\% & 0.36 & 4.00\% & 34.00\% & 0.22 & 3.00\% & 31.00\% \\
LLMGenPlan ~\cite{silver2024generalized} & 1.24 & 23.00\% & 61.33\% & 1.83 & 42.00\% & 70.00\% & 1.14 & 14.00\% & 67.00\% & 0.76 & 13.00\% & 47.00\% \\
DELTA~\cite{liu2025DELTA} & 2.25 & 46.33\% & 67.00\% & 2.86 & 69.00\% & 74.00\% & 1.94 & 31.00\% & 68.00\% & 1.94 & 39.00\% & 59.00\% \\
\midrule
\textbf{SCOPE} & \textbf{3.28} & \textbf{70.67\%} & \textbf{92.67\%} & \textbf{3.58} & \textbf{86.00\%} & \textbf{93.00\%} & \textbf{3.04} & \textbf{55.00\%} & \textbf{94.00\%} & \textbf{3.22} & \textbf{71.00\%} & \textbf{91.00\%} \\
\bottomrule
\end{tabular}%
}
\end{table*}
\begin{table*}[t]
\vspace{-4pt}
\centering
\caption{Generalization Performance of Fine-tuned Models on AI2-THOR Grounded Dataset. Ablation of the planning capabilities of models fine-tuned on datasets varying in scene diversity and error-correction.}
\vspace{-5pt}
\label{tab:dataset_b_results}
\resizebox{\textwidth}{!}{%
\begin{tabular}{lcccccccccccccc}
\toprule
\multirow{2}{*}{\textbf{Methods}}& \multirow{2}{*}{\textbf{Div.}} & \multirow{2}{*}{\textbf{Corr.}} & \multicolumn{3}{c}{\textbf{Overall}} & \multicolumn{3}{c}{\textbf{Simple Tasks}}& \multicolumn{3}{c}{\textbf{Complex Tasks}} & \multicolumn{3}{c}{\textbf{Vague Tasks}} \\
\cmidrule(lr){4-6} \cmidrule(lr){7-9} \cmidrule(lr){10-12} \cmidrule(lr){13-15}
 & & & \textbf{AQS~$\uparrow$} & \textbf{SR~$\uparrow$} & \textbf{EX~$\uparrow$} & \textbf{AQS~$\uparrow$} & \textbf{SR~$\uparrow$} & \textbf{EX~$\uparrow$} & \textbf{AQS~$\uparrow$} & \textbf{SR~$\uparrow$} & \textbf{EX~$\uparrow$} & \textbf{AQS~$\uparrow$} & \textbf{SR~$\uparrow$} & \textbf{EX~$\uparrow$} \\
\midrule
LLM-As-Planner~\cite{song2023llm} & \textcolor{Dred}{\ding{55}} & \textcolor{Dred}{\ding{55}} & 1.57 & 25.00\% & 26.00\% & 2.43 & 54.00\% & 54.00\% & 1.36 & 10.00\% & 11.00\% & 0.91 & 11.00\% & 13.00\% \\
DELTA \cite{liu2025DELTA} & \textcolor{Dred}{\ding{55}} & \textcolor{Dred}{\ding{55}} & 2.67 & 50.67\% & 77.67\% & 3.47 & 83.00\% & 90.00\% & 2.23 & 28.00\% & 74.00\% & 2.31 & 41.00\% & 69.00\% \\
\midrule
SCOPE & \textcolor{Dgreen}{\ding{51}} & \textcolor{Dred}{\ding{55}} & 3.03 & 59.67\% & 87.00\% & 3.62 & 87.00\% & 93.00\% & 2.66 & 40.00\% & 86.00\% & 2.81 & 52.00\% & 82.00\% \\
SCOPE & \textcolor{Dred}{\ding{55}} & \textcolor{Dgreen}{\ding{51}} & 2.96 & 58.00\% & 81.00\% & 3.60 & 86.00\% & 93.00\% & 2.65 & 41.00\% & 71.00\% & 2.64 & 47.00\% & 79.00\% \\
\textbf{SCOPE} & \textcolor{Dgreen}{\ding{51}} & \textcolor{Dgreen}{\ding{51}} & \textbf{3.08} & \textbf{63.00\%} & \textbf{85.33\%} & \textbf{3.61} & \textbf{88.00\%} & \textbf{93.00\%} & \textbf{2.83} & \textbf{50.00\%} & \textbf{83.00\%} & \textbf{2.81} & \textbf{51.00\%} & \textbf{80.00\%} \\
\bottomrule
\vspace{-19pt}
\end{tabular}%
}
\end{table*}
\subsubsection{Evaluation Metrics and Baselines} 

To evaluate embodied planning performance, we propose a physically-grounded semantic evaluation framework. We utilize PDDLGym and VAL toolchains to verify action sequence physical executability. Subsequently, an LLM-as-a-Judge approach assesses valid physical simulation trajectories, defining three core metrics: \textbf{Executability (EX)} measures the fraction of successfully executed environmental actions, regardless of overall task relevance; \textbf{Average Quality Score (AQS)} employs a 0--4 fine-grained grading system covering dimensions from complete failure, partial completion, suboptimal success (redundancy or implicit omissions) to perfect execution; and \textbf{Success Rate (SR)} counts only perfect-scoring tasks (Score = 4) as successful, strictly penalizing logical flaws to ensure results reflect real-world high availability.

To comprehensively validate SCOPE, we select four representative baselines for comparison:

\begin{itemize}
    \item \textbf{LLM-As-Planner}\cite{song2023llm}: An open-loop planning paradigm directly concatenating environmental states and goals into a prompt, querying the LLM to generate high-level action sequences in a single pass.
    \item \textbf{LLM+P}\cite{liu2023llm}: A neuro-symbolic method utilizing LLMs to translate natural language instructions into PDDL problem files, invoking a classical automated planner for precise solving.
    \item \textbf{LLMGenPlan}\cite{silver2024generalized}: An open-loop approach summarizing domain knowledge into executable Python code, generating task plans by executing it.
    \item \textbf{DELTA}\cite{liu2025DELTA}: A hierarchical planning method auto-regressively decomposing long-horizon tasks into subgoal sequences delegated to a PDDL solver for step-by-step execution.
\end{itemize}

\begin{figure*}[t!]
    \centering
    \vspace{3pt}
    \includegraphics[width=\textwidth]{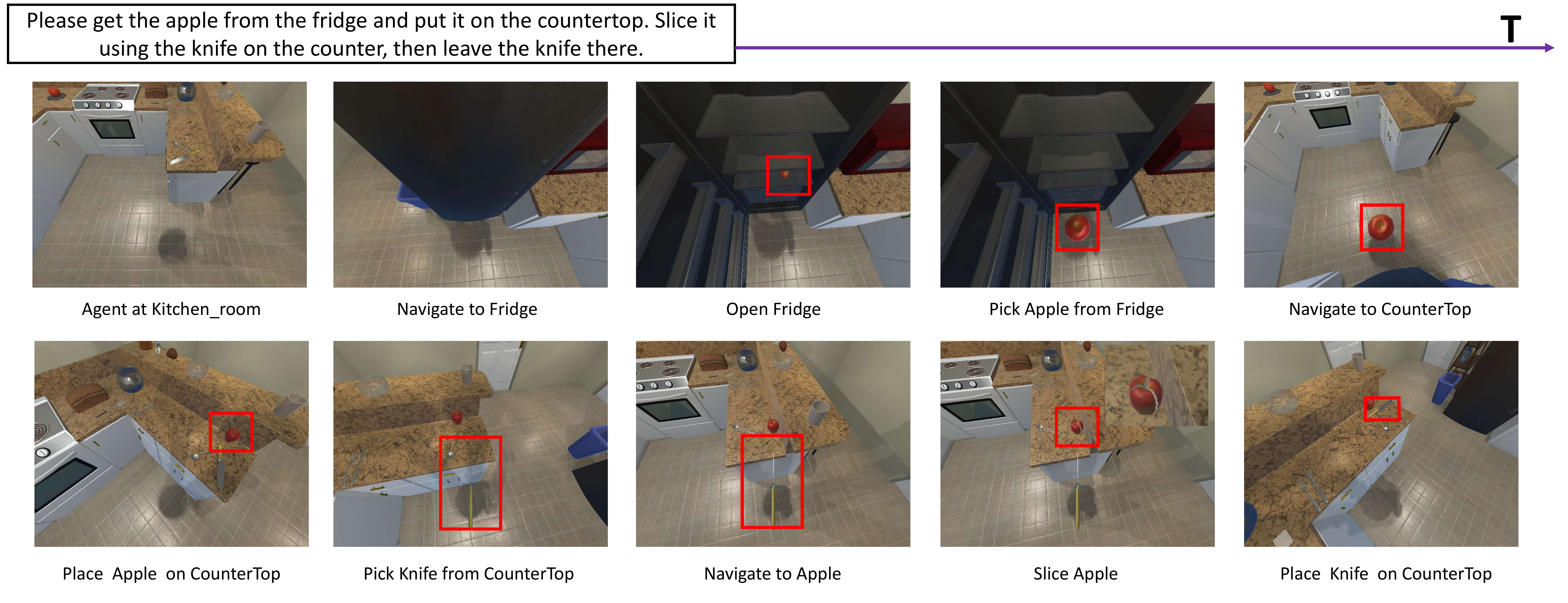}
    \vspace{-20pt}
    \caption{Simulation snapshots of the execution process of a long-horizon task.}
    \label{fig:ai2thor_viz}
    \vspace{-13pt}
\end{figure*}

\subsection{Planning Robustness and Repair Capabilities under Erroneous Data}
\label{subsec:robustness_repair}

In this evaluation, all planning methods directly receive the same noisy scene graphs and task instructions, and no model fine-tuning is involved. The evaluation results are presented in Tab.~\ref{tab:robustness_dataset_a}. Among all methods, LLMGenPlan exhibits the poorest performance, achieving an overall success rate of merely 6.33\% and an average score as low as 0.40. The low executability rate of 36.67\% confirms that although LLMs possess a certain degree of temporal reasoning capability, relying directly on their internal weights for prediction is fundamentally a probabilistic token generation process rather than a rigorous search and optimization procedure. The introduction of a retry mechanism in LLMGenPlan Replan improves the success rate on simple tasks from 12.00\% to 42.00\% and significantly boosts the overall executability to 61.33\%. However, the substantial gap between its executability (61.33\%) and success rate (23.00\%) indicates that this text-validation-based feedback only filters out invalid action expressions and fails to rectify logical errors in the underlying scene graph, resulting in plans that are physically executable but semantically incorrect. When confronted with complex tasks involving long-horizon dependencies, its success rate drops to a mere 14.00\%. Similarly, LLM-As-Planner, which adopts a fundamental prompting paradigm, generates plans with an extremely low executability of 29.67\% when facing noisy scene graphs with missing attributes, resulting in an overall success rate of only 27.00\%.

By translating natural language instructions into PDDL, LLM+P and DELTA delegate the planning process to automated task planners. The critical flaw in both methods lies in their absolute rigid dependency on the initial state. In LLM-Synthetic Noisy Dataset, the scene graphs inevitably contain logical hallucinations and attribute omissions. Consequently, classical PDDL solvers fail outright due to unsatisfied preconditions, returning no solution or triggering a timeout. This passive discard-on-error strategy limits the executability of generated plans to 48.00\% and 67.00\% respectively, causing their average scores on complex tasks to stagnate below 1.94, with success rates of only 30.00\% and 31.00\%.

The SCOPE framework achieves comprehensive superiority, attaining a high overall success rate of 70.67\%, a remarkable executability rate of 92.67\%, and an average score of 3.28. Particularly in complex and vague tasks, SCOPE secures exceptionally high success rates of 55.00\% and 71.00\%, respectively, significantly outperforming the strongest baseline, DELTA (an increase of over 24\%). When SCOPE's solver halts due to missing states, it dynamically reverse-traces the missing preconditions to complete and repair the underlying scene graph. Through this closed-loop planning-feedback-repair mechanism, SCOPE not only shatters the limitations of traditional formal solvers but also ensures that the generated plans are physically grounded, achieving near-perfect executability (94.00\% on complex tasks), ultimately yielding high-quality data.

\subsection{Impact of Scene Diversity and Error-correction on Model Fine-Tuning}
\label{subsubsec:generalization_eval}

Tab.~\ref{tab:dataset_b_results} presents the evaluation results of the fine-tuned models on the \textbf{AI2-THOR Grounded Dataset}, analyzing the effects of scene diversity and error correction on model generalization. We adopt \textbf{Qwen2.5-7B-Instruct} as the foundation model. We first construct two types of scene graph datasets: conventional scenes based on high-frequency preferences (\textit{w/o Div.}) and diverse scenes generated using entropy-aware inverse-frequency weighting (\textit{w Div.}). Subsequently, we use LLM-as-Planner, DELTA, and different ablation configurations of SCOPE to generate separate training datasets, each containing 10,000 $\langle SG_{\mathrm{final}}, I, \Pi \rangle$ triplets. Finally, we fine-tune the foundation model on each dataset using identical LoRA settings and evaluate the downstream planning performance of the resulting models on the \textbf{AI2-THOR Grounded Dataset}.
\subsubsection{Impact of Scene Diversity on Fine-Tuning Performance}
Comparing the impact of data generated by different planning models on TAMP downstream performance, introducing diverse scene generation (\textit{w Div.}) increases model performance, confirming the critical importance of covering long-tail distributions to shatter path dependency. DELTA's success rate increases from 50.67\% to 59.67\% alongside an executability boost to 87.00\%, with the complex task success rate increasing by nearly 12\%, indicating diverse environmental data effectively enhances adaptability and physical compliance in complex topologies.

\subsubsection{Impact of Error-correction on Fine-Tuning Performance}
Restricting the training data source to conventional scenes (\textit{w/o Div.}) ablates the impact of the plan feedback error-correction. The model fine-tuned on LLM-as-Planner generated data exhibits the poorest performance, yielding an overall success rate of 25.00\% with 26.00\% executability. This high correlation between failure and physical invalidity reflects that erroneous open-loop LLM data severely misguide the model's physical reasoning capabilities. In contrast, fine-tuning with DELTA generated data significantly elevates the success rate to 50.67\%. Remarkably, the model fine-tuned with SCOPE-generated data achieves the highest overall success rate of 58.00\% and 81.00\% executability under the identical setup. Particularly in complex tasks, its 41.00\% performance significantly outpaces DELTA's 28.00\%. These results substantiate that SCOPE's plan feedback error-correction (\textit{w Cor.}) acts as a data cleanser, rectifying initially inexecutable hallucinated trajectories into high-quality training samples strictly adhering to physical constraints. Even without increasing scene diversity, the high-quality plans provided by SCOPE enhance LLM planning performance.

Meanwhile, by integrating the error-correction with the environmental diversity from entropy-aware generation, SCOPE achieves the highest overall success rate of 63.00\%, an executability of 85.33\%, and an average score of 3.08.

Furthermore, to evaluate interactive environment performance, as illustrated in Fig. \ref{fig:ai2thor_viz}, we deploy the plans generated by the SCOPE-fine-tuned model into the AI2-THOR simulator for execution validation. Fig. \ref{fig:ai2thor_viz} shows the model accurately generates action sequences compliant with physical constraints and successfully accomplishes long-horizon tasks involving multi-step operations.
\begin{figure}[htpb]
    \vspace{3pt}
    \centering
    \includegraphics[width=0.95\columnwidth]{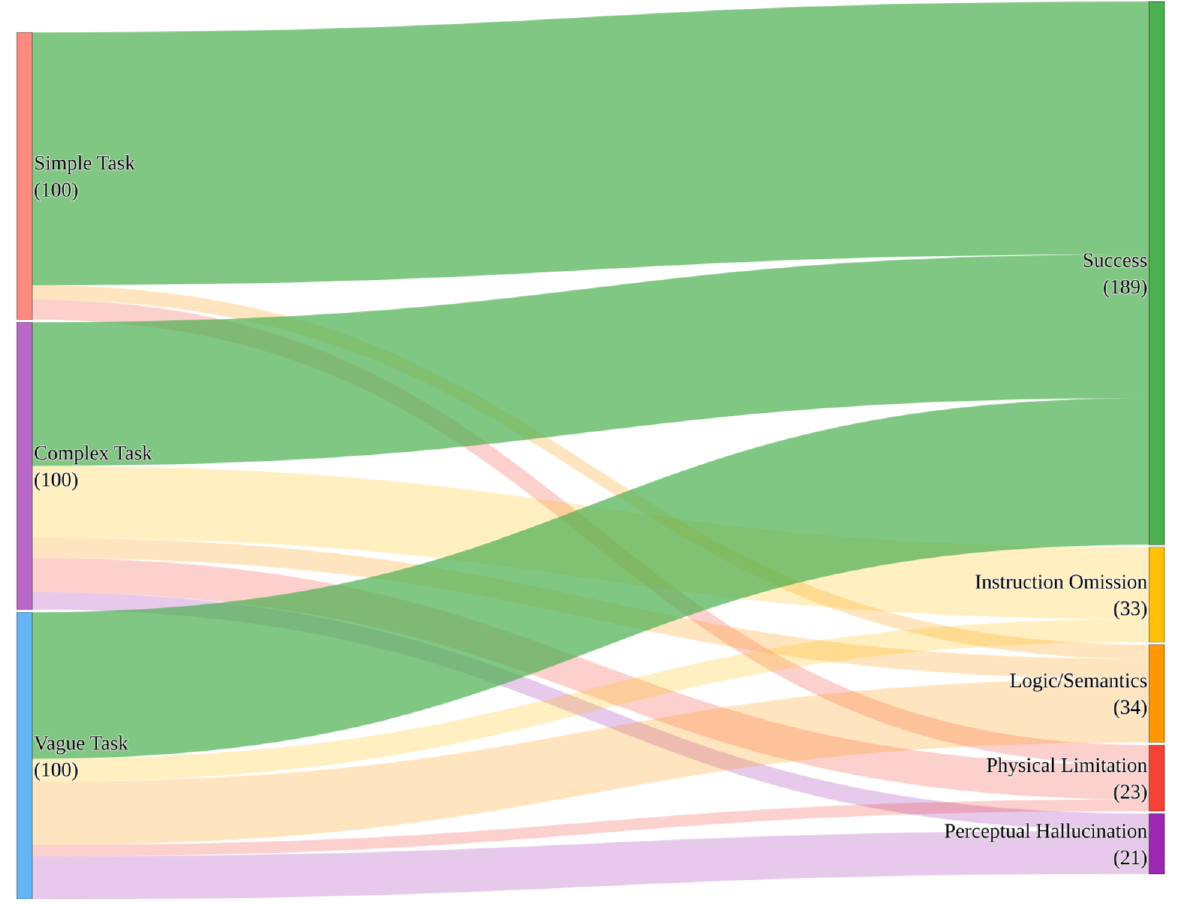}
    \vspace{-5pt}
    \caption{Failure analysis of the SCOPE-fine-tuned model. The errors are classified into high-level cognitive deficits and low-level grounding issues to illustrate the bottlenecks in the planning pipeline.}
    \label{fig:error}
    \vspace{-15pt}
\end{figure}

Despite these improvements, limitations remain. As shown in Fig.~\ref{fig:error}, we classify the 111 failed cases into high-level cognitive deficits and low-level grounding issues. Cognitive errors account for about 60\% of failures, including instruction omission (33 cases) and logic and semantic errors (34 cases). The remaining failures arise from physical limitations (23 cases), such as violating environmental physics, and perceptual hallucinations (21 cases), such as interacting with non-existent objects.
\section{Conclusion}
In this work, we introduced SCOPE, an automated framework for generating diverse and physically executable datasets for long-horizon robot planning. By integrating entropy-aware scene generation with plan feedback error-correction, SCOPE effectively addresses  challenges of homogeneity and logical hallucinations in LLM-based data generation. Experiments demonstrate that SCOPE not only achieves stronger planning robustness under noisy conditions but also serves as a high-quality data engine; models fine-tuned on SCOPE-generated data exhibit substantial generalization improvements in realistic simulations, highlighting the value of closed-loop data synthesis for advancing embodied intelligence. Future work will target more complex, dynamic environments.

\section*{ACKNOWLEDGMENT}
This work was supported by the Strategic Priority Research Program of the Chinese Academy of Sciences under Grants XDA045000 and XDA0450202, and by the Central Government Guidance Funds for Local Science and Technology Development under Grant YDZX2025124.

\bibliographystyle{IEEEtran}
\bibliography{main}{}
 
\end{document}